# SIMILARITY MEASURING APPROACH FOR ENGINEERING MATERIALS SELECTION

**DORESWAMY[1], M.N. VANAJAKSHI[2]**

[1]*Department of Post-Graduate Studies and Research in Computer Science*
*Mangalore University, Mangalagangotri-574 199, Karnataka, INDIA*
*Phone:+91-824-2287670, E-mail:* doreswamyh@yahoo.com

[2] *Department of Computer Science and Engineering*
*PES College of Engineering, Mandya, INDIA.*
*E-mail:* vanaja.sp@gmail.com

## Abstract

Advanced engineering materials design involves the exploration of massive multidimensional feature spaces, the correlation of materials properties and the processing parameters derived from disparate sources. The search for alternative materials or processing property strategies, whether through analytical, experimental or simulation approaches, has been a slow and arduous task, punctuated by infrequent and often expected discoveries. A few systematic efforts have been made to analyze the trends in data as a basis for classifications and predictions. This is particularly due to the lack of large amounts of organized data and more importantly the challenging of shifting through them in a timely and efficient manner. The application of recent advances in Data Mining on materials informatics is the state of art of computational and experimental approaches for materials discovery. In this paper similarity based engineering materials selection model is proposed and implemented to select engineering materials based on the composite materials constraints. The result reviewed from this model is sustainable for effective decision making in advanced engineering materials design applications.

Keywords: Data Mining and Knowledge Discovery, Composite Materials Selection, Similarity Measure.

## 1. Introduction

Engineering materials are the artificial materials, such as Polymer, Ceramic, Metal and their composite with fiber reinforced materials, which are being used in our daily life. Any two materials could be combined to make a composite and they might be mixed in much geometry. Selection of design and fabrication processes associated to engineering materials design is the tedious task that is being faced by the most of the manufacturing industries. The selection of appropriate materials, which meet the design requirements and improve the performance, reliability, durability of composite material, is the critical task in Computer Aided Design (CAD) and Computer Aided Manufacturing (CAM) systems[5].

As wide variety of more than 50000 materials available today and varying in their characteristics and costs, materials selection system is very much essential to ease the difficult complex process. This selection process involves decision-making strategies in determining the prerequisite materials that suit the design specifications and requirements of composite design.

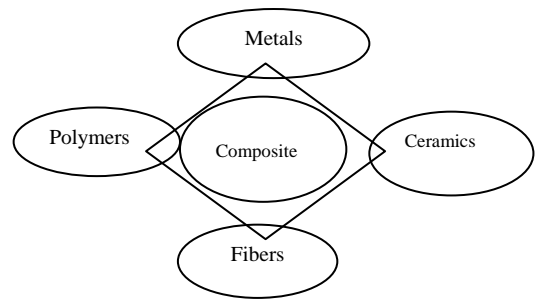

Fig. 1: The material classes from which composite are made

Cost effective materials selection that meet the design requirements reduces the manufacturing cost, increases manufacturing throughput and reduces the materials selection complexity as posed by a designer. To automate computer aided manufacturing systems, various intelligent decision support systems were designed [1] [2] [3] [4] [8] [11] [12]. The applications of expert system play major role in diverse application fields from materials design and their manufacturing. Design of computational expert systems on wider range of data sets have still research scope in advanced engineering materials design applications[6][13][14]. Therefore, Composite Materials Selection System (CMSS) is proposed and implemented in this paper.

The paper has been organized as follows. The second section presents the composite materials selection system. The third section describes similarity measure functions. The forth section describes the selection strategy on different materials type. The last section concludes the work and briefs the future work scope.

**2. Composite Materials Selection System**

Expert systems are programs in which domain knowledge about a problem is embedded in a set of modules called as rules, frames, objects, or scripts that are stored in a repository called a knowledgebase. The Composite Materials Selection System (CMSS) is developed in order to simplify the complex selection process for opting appropriate materials that meet the design requirements. The structure of the proposed system is shown in the figure 2.

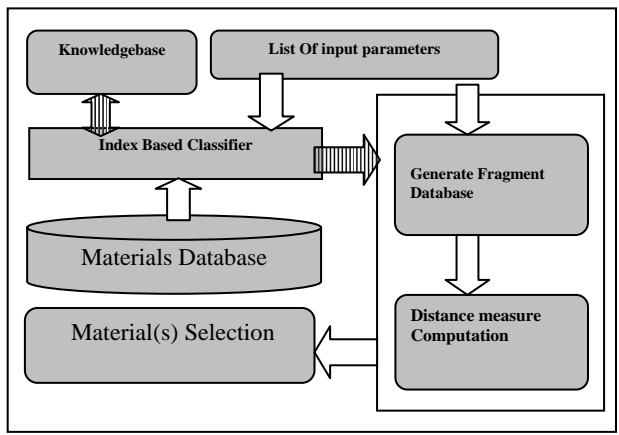

Fig. 2. Composite Material Selection System (CMSS)

The CMSS consists of several integrated modules that are responding for potential input parameters to produce outputs that are treated as inputs of another module. The integrated modules of CMSS are input module, Indexed based classifier [9] [10], fragment database generator, distance measure computation module and materials selection module. All these modules are simplified with non-redundant computational effort.

The input module (list of input parameters) provides the CMSS a list of materials characteristics that are specified by design engineers. It will be interacting with both the indexed based classifier and fragment database generator. Index based decision classifier scans through the inputs and segregates materials characteristics/ attributes into different classes that are represented by nodes. The segregation of attributes into different classes based on the classification rules defined in the knowledgebase of the system. The outcome of index based decision classifier is forwarded to the fragment database generator that selects the portion of the database containing matching attributes with the tuples belonging to materials class as predicted by the index classifier.

*2.1. Composite Design Specifications*

The composite design specifications are the parameters [6] of a component to be designed and a design engineer derives these parameters. Design requirements are the properties of primary importance such as physical properties, mechanical properties, chemical properties, thermal properties and so on. These properties represent quantitative attribute and linguistic values of a component. There are 23 properties considered in this system. Some quantitative properties have range values (Density: $0.23cm^3$ to $0.56cm^3$) and others properties have ordinals/linguistic/categorical values (Poor- Excellent). Each ordinal/linguistic value is replaced with a unique numeric weight.

*2.2. Composite Material Database organization*

Material Database(MD) consists of different classes of materials such as Polymer, Ceramic and Metal. All materials are having the same set of properties but some of them are linguistic properties.

## 2.3. Knowledgebase

Knowledgebase [7] is defined as "A database of knowledge about a subject; used in Artificial Intelligence. The knowledgebase for an expert system (a computer system that solves problems) comes partly from human experience and partly from the computer's experience in solving problems. It must be expressed in a formal knowledge representation language for the computer to use it". The knowledgebase of CMSS consists of 23 decision rules and each decision rule generates a prime index pattern that represents a material class.

## 2.4. Index Based Classifier

Index based decision classier shown in figure 3 is a simple and robust classifier that is used as decision-making principles in most of the fields such as Machine Learning, Pattern Recognition, Image Processing and Data Mining and Knowledge Discovery. It discriminates design requirement properties based on the expert rules defined in the knowledgebase. Each class generated by the classifier is implemented with linked lists. Each node in a list has three fields including Property Name (PN), Property Index(PI) and a Pointer(Ptr) for respectively storing the next property name as defined in the input design requirement list, index value generated by the decision classifier, and the next node address.

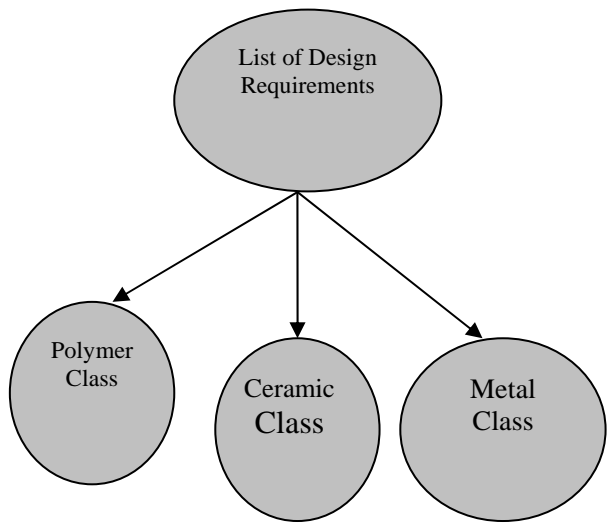

Fig. 3: Index Based Classifier

In the first step of classifier, when a property is randomly sampled from a design requirement list, the classifier invokes the rules defined in the knowledgebase and creates a node in the class corresponding to the index pattern.

## 2.5. Generating Fragment Database

The Material Database (DM) stores all classes of materials, C = {P, C, M}. Each class is having the materials attributes, $A = \{a_1, a_2, a_3, a_4, a_5, \ldots \ldots a_m\}$. A class of materials fragmented from material database is proportional to O(N) time complexity in the best, average and worst cases of analysis. This fragmented data space reduces the computational efforts with less memory space during computing distance measure values.

A Fragmented Database(FD) consists of $N$ number of tuples, $T = \{t_1, t_2, t_3, t_4, t_5, \ldots \ldots t_N\}$, each tuple, $t_i$, consists of $m$ materials attributes, $A = \{a_1, a_2, a_3, a_4, a_5, \ldots \ldots a_m\}$. The design requirements of class $C_i$ represented by the set $R = \{r_1, r_2, r_3, r_4, r_5, \ldots \ldots r_n\}$ are the properties specified by the design engineers. The unwanted attributes in the tuples of FD can be eliminated still for reducing the computational complexity. The relevant design requirement properties of interest are obtained by the following set operation.

$$FD_n = R \cap A, \ R \in A \quad (1)$$

The resultant database obtained by (1) is represented with object – by- variable structure. The structure in the form of a relational table of N-by-n is a data matrix and is represented as bellow:

$$D_{N \times n} = \begin{bmatrix} x_{1,1} \ldots \ldots x_{1,f} \ldots \ldots x_{1,n} \\ \ldots \ldots \quad \ldots \quad \ldots \\ x_{i,1} \ldots \ldots x_{i,f} \ldots \ldots x_{i,n} \\ \ldots \ldots \quad \ldots \quad \ldots \\ x_{N,1} \ldots \ldots x_{N,f} \ldots \ldots x_{N,n} \end{bmatrix} \quad (2)$$

Doreswamy

The minimum distance between the input data set $y = \{y_1, y_2, y_3, y_4, y_5, \ldots\ldots y_n\}$ and a feature set $x = \{x_1, x_2, x_3, x_4, x_5, \ldots\ldots x_n\}$ in the data space, $D_{N \times n}$, is computed using a distance measure functions.

## 3. Distance Measure Computation

Similarity/Distance measure functions are used to compute the logical distance between the input data set say, *y,* and the data set, *x,* in a data space. The applications of these are employed in Data Mining and Knowledge Discovery fields [7] for data classification and clustering analysis. Any function is said to be distance metric function if it satisfy all the four conditions(1-4), otherwise similarity function if it satisfies the first three following conditions:

1. $d(x,y) \geq 0$; the distance is a non-negative number.

2. $d(x,x) = 0$; the distance of an object to itself is zero.

3. $d(x,y) = d(y,x)$; The distance is symmetric function

4. $d(x,z) + d(z,y) \geq d(x,y)$; Going directly from an object, x, to an object, y, in space is no more than making a detour over any other object other than object z (triangular inequality).

There are various popular distance measuring functions that are satisfying the above principles. Euclidian distance measure [7] metric is employed for distance computations. This distance measure metric is defined as follow:

$$d(y,x) = \sqrt{\sum_{i=1}^{n}(y_i - x_i)^2} \quad (3)$$

where $x = \{x_1, x_2, x_3, x_4, x_5, \ldots\ldots x_n\}$ and $y = \{y_1, y_2, y_3, y_4, y_5, \ldots\ldots y_n\}$ are two *n* dimensional data objects.

### i. City Block Distance Metric

$$d(y,x) = \sum |y_i - x_i| \quad (4)$$

### iii. Absolute Exponential measure:

$$d(y,x) = \exp\left(-\sum_{k=1}^{n}|y_k - x_k|\right) \quad (5)$$

### iv. Geometric Average Minimum:

$$d(y,x) = \frac{\sum_{k=1}^{n} \min(x_k, y_k)}{\sum_{k=1}^{n}[x_k \cdot y_k]^{1/2}} \quad (6)$$

### v. Correlation Coefficient measure:

$$d(y,x) = \frac{\sum_{k=1}^{n}|x_k - \bar{x}_i||y_k - \bar{y}_j|}{\left[\sum_{k=1}^{n}(x_k - \bar{x}_j)^2\right]^{1/2}\left[\sum_{k=1}^{n}(y_k - \bar{y}_j)^2\right]^{1/2}} \quad (7)$$

$$\bar{x}_i = \frac{1}{n}\sum_{k=1}^{n}x_k, \quad \bar{y}_j = \frac{1}{n}\sum_{k=1}^{n}y_k$$

### vi. Exponential Similarity Measure:

$$d(y,x) = \sum_{k=1}^{n}\frac{|y_k - x_k|}{1 + e^{-|y_k - x_k|}} \quad (8)$$

where $x = \{x_1, x_2, x_3, x_4, x_5, \ldots\ldots x_n\}$ and $y = \{y_1, y_2, y_3, y_4, y_5, \ldots\ldots y_n\}$ are two *n* dimensional data objects.

## 4. Similarity Material Selection

It is the process of selecting the best match data object in data space for an input data object. The best match for an input data object is determined by the Euclidian distance computation. This has been using as standard distance measure function in data mining and knowledge discovery [7]. The best match object for an input object is selected through the determination of the least similarity measure value.

$$\min\{d(i)\}_{i=1}^{N} = \sum_{j=1}^{n}|y_j - D_{i,j}| \quad (9)$$

## 5. Experimental Simulation Results

Material database used in the CMSS has 2000 materials data sets including Polymers Ceramics and Metals. Each one has 23 properties and includes both numerical and categorical values. The categorical values have been predefined with numeric values so that to distinguish them numerically among them in the computations. The Euclidian distance between the input requirements and the properties of each material in the

Similarity Measuring Approach For

selected class are computed. A material corresponding to the least distance is selected as the potential material that meets the design requirements.

Design specification specified by design engineers are the input parameters that enabled the CMSS through the form shown in figure 4. Initial classification on input design requirements into Polymer, Ceramic and Metal Classes, the fragment materials data sets generated and the material selected by the Euclidian measuring technique from the different classes are shown in the figure 5.

Fig. 4: Input Design requirement form

Fig. 6: Materials selected using Euclidian Distance Measure.

Fig. 5: Fragmented materials and the materials selected from the respective selected materials data set.

Doreswamy

Fig. 6(a): Materials selected using City-Block Distance Measure.

Fig. 6(b): Materials selected using Absolute Exponential Measure.

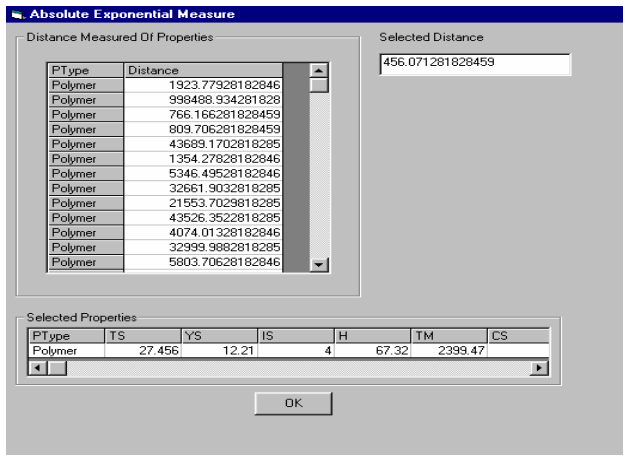

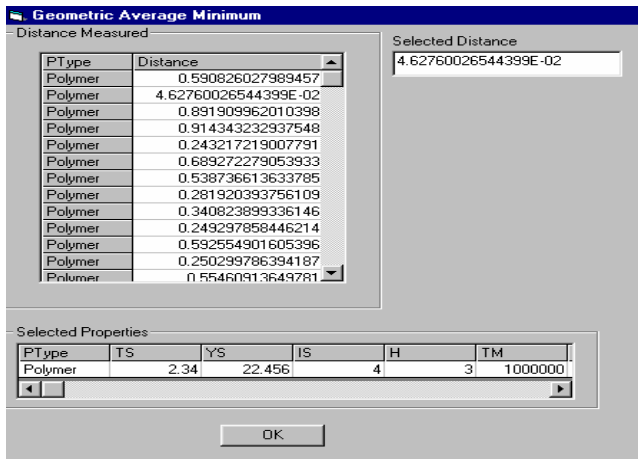

Fig. 6(c): Materials selected using Geometric Average Measure.

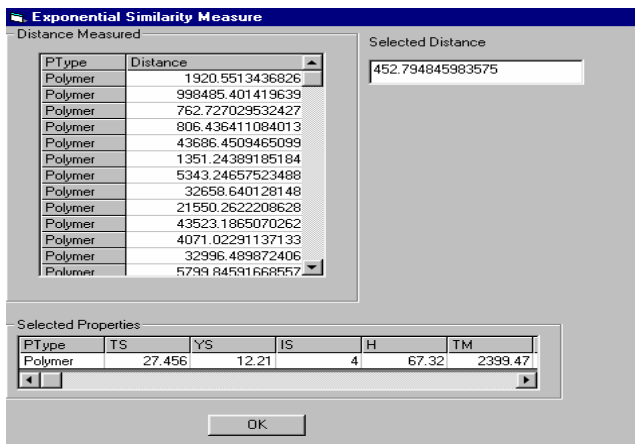

Fig. 6(d). Materials selected using Exponential Similarity Measure.

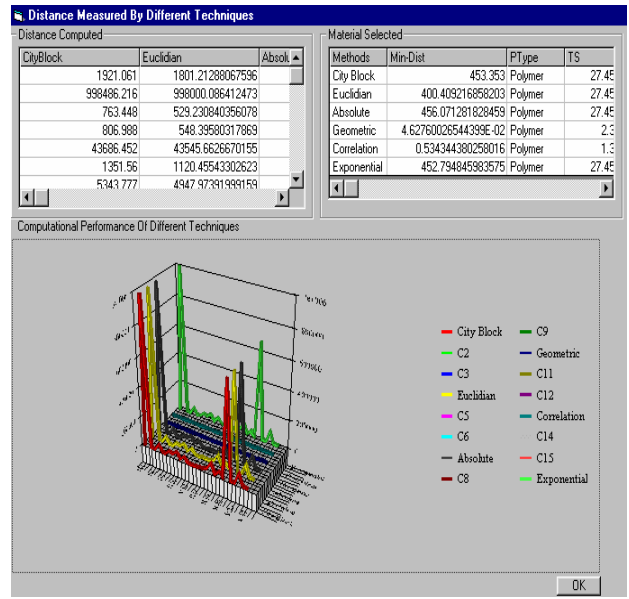

Fig. 7: Performance of evaluation of different Distance measure Techniques.

Different distance measure computations performed for materials selection are shown in figures 6-6(d). Polymer class properties generated by the indexed classifier are listed in the table 1. Materials selected over the degree of similarity computed between the properties in table 1 and in the fragmented data sets are shown in the table 2. From this table 2, it shows that distance/similarity measure functions (3), (4), (5), (7) and (8) belonging to $L_1$ family and are competent enough to select the materials that are very closure to the input specification. However, the function (6) and (7) are feasible for materials selection, but function (7) is more appropriate for analyzing redundancy and consistency among the materials data sets. Function (6) is not the feasible one as it maps to the different material in the class.

The $L_1$ family functions and the functions (6) and (7) are compared and shown in the Table 2 and their performance evaluation on numeric approximation is depicted in the figure 7. The degree of similarity of Euclidian distance function is less that emphasizes much closeness among the $L_1$ family functions. The degrees of similarity of functions (6) and (7) depicted in the table 2 ( Sl. nos. 4 and 5 ) are still less than the Euclidian distance measure, however, one of these functions(6) maps the input design requirements to the materials that do not guarantee the optimal and expected design requirements performance.

Table 1: Input parameter list associated to a material under Polymer class(Y)

| Tensile Strength | Yield Strength | Impact Strength | Hardness | Tensile Modulus |
|---|---|---|---|---|
| 20.00 | 23.90 | 4.00 | 56.67 | 2000.00 |

Table 2: Degree of similarity and materials selected from the Polymer class by different distance measuring functions

| Sl.No. | Distance / Similarity Measure Functions | Degree of Similarity | Materials Selected from the Polymer Class | | | | |
|---|---|---|---|---|---|---|---|
| | | | Tensile Strength | Yield Strength | Impact Strength | Hardness | Tensile Modulus |
| 1 | Euclidian Distance Measure | 400.4097 | 27.456 | 12.21 | 4 | 67.32 | 2399.47 |
| 2 | City Block Distance Measure | 453.353 | 27.456 | 12.21 | 4 | 67.32 | 2399.47 |
| 3 | Absolute Exponential Measure | 456.071 | 27.456 | 12.21 | 4 | 67.32 | 2399.47 |
| 4 | Geometric Average Measure | 4.6270 | 2.34 | 22.456 | 4 | 3 | 1.0E+06 |
| 5 | Correlation Coefficient Measure | 0.89343 | 27.456 | 12.21 | 4 | 67.32 | 2399.47 |
| 6 | Exponential Similarity Measure | 452.7948 | 27.456 | 12.21 | 4 | 67.32 | 2399.47 |

## 6. Conclusion And Future Work Scope

Effective design of materials and their composites includes complex redundant computational efforts. These redundant computational efforts are reduced in the MCSS. Simple and robust Fragment Database(FD) was generated for speeding up the selection processes and for removing materials attributes that were not consistent for measuring similarity between two materials. Euclidian distance measuring function is compared with exponential similarity measure function[11] that approximates the similarity value than the city block distance measure values in eliminating the outliers from the large data set. One of the disadvantages of the Euclidian distance metric function is that if one of the input attributes has a relatively large range, then it can overpower the other attributes. For example if an application has just two attributes, X and Y, and X can have values from 100.0 to 1000.0, and Y can have values only from 10.0 to 100.0 then Y's influence on the distance function will usually be overpowered by X's influence. Therefore, distance are required to be normalized by dividing the distance for each attribute by the range that is maximum-minimum of that attribute so normalize to desired range. This family of distance measures are not suitable for ordinal values.

The CMSS would be failure when the attributes values are very small and majority of the attributes having categorical values. This declines the selection performance. This drawback of this system can be overridden through the supervised learning neural

Doreswamy

network algorithm with fuzzy based axioms for approximating categorical and numerical values.

Further, this module can also be extended as an effective decision support system for extracting relevant knowledge of materials and their properties for designing high performance composite materials.

**Acknowledgement**

This work has been supported by the University Grant Commission (UGC), India under Major Research Project entitled "Scientific Knowledge Discovery Systems (SKDS) For Advanced Engineering Materials Design Applications" vide reference no. 34-99\2008 (SR), 30th December 2008. The author gratefully acknowledges the support.

**References**

[1]. Wenham Zhang, Michael J. Bazooka, Laager Kari and Eric J. Amiss, An open source Informatics Systems for Combinatorial Materials Research, Polymeric Materials: Science & Engineering, 2004, 90,341.

[2]. Krishna Rajan and Mohammed Sake, "Data Mining through Information Association: A Knowledge Discovery tool for Materials Science", *CODATA proceedings*, Beaver, Italy, 2002.

[3]. Ronald E. Giachetti," Decision Support System for Material and Manufacturing Selection" *Journal of Intelligent Manufacturing, January,* 1997, (5): 656-671.

[4]. D. Bourell, "Decision Matrices in Material Selection"., ASM metals hand book, Vol. 20, Volume Chair George Dieter, ASM International, Materials Park,OH,1997, pp.243-254.

[5]. P.A. Gutteridge and J. Turner," Computer Aided Materials Selection and Design", Journal of *Materials and Design*, 3, August 1982, pp.504-510.

[6]. Michael Goebel and Le Greenwood, "A Survey Of Data Mining And Knowledge Discovery Software Tools", *ACM SIGKDD,* June 1999, Vol.1 (1): 20-32.

[7]. Jawed Han and Michelin Camber, "Data Mining Concepts and Techniques", Elsewhere Science, India 2002.

[8]. Tokyo, McGraw-Hill, Edwards, K, L. "Towards more effective decision support in materials and Design Engineering". *Materials and design*: 1994. 5(5):251-258.

[9]. Doreswamy, S. C. Sharma, and M Krishna, "Knowledge Discovery System for Cost-Effective Composite Polymer Selection-Data Mining Approach", *12th International Conference on Management of Data COMAD 2005b*, Hyderabad, India, December 20-22, 2005, pp. 185-190.

[10]. Doreswamy, S. C. Sharma, M. Krishna and H N Murthy, "Data Mining Application In Knowledge Extraction Of Polymer And Reinforcement Clustering", *Proceedings of International Conference on Systemic, Cybernetics And Informatics*, Pentagram Research Center, Hyderabad, INDIA, January 4th to 8$^{th}$ , 2006, pp.562-566.

[11]. Doreswamy and S.C. Sharma, "An Expert Decision Support System for Engineering Materials Selections And Their Performance Classifications on Design Parameters", *International Journal of Computing and Applications (ICJA)*, June 2006. Vol.1 (1):17-34.

[12]. Doreswamy, "Engineering Materials Classification Model- A Neural Network Application", *IJDCDIS A Supplement, Advances in Neural Networks*, 2007, Vol. (14) (S1): 591-595.

[13]. Doreswamy, "A survey for Data Mining framework for polymer Matrix Composite Engineering materials Design Applications" *International Journal of Computational Intelligence Systems (IJCIS)*, 2008. Vol.1(4): 312-328,

[14]. Krishna Rajan, Materials Informatics Part 1; A diversity of Issues, Journal of Materials, March 2008.